\documentclass{article}

\usepackage{microtype}
\usepackage{graphicx}
\usepackage{booktabs} % for professional tables

\usepackage{hyperref}

\usepackage[accepted]{icml2025}

\usepackage{amsmath}
\usepackage{amssymb}
\usepackage{mathtools}
\usepackage{amsthm}
\usepackage{subcaption}
\usepackage{multirow}

\usepackage[capitalize,noabbrev]{cleveref}

\theoremstyle{plain}

\theoremstyle{definition}

\theoremstyle{remark}

\usepackage[textsize=tiny]{todonotes}

\icmltitlerunning{Random-Set Large Language Models}

\begin{document}

\twocolumn[
\icmltitle{Random-Set Large Language Models}

% It is OKAY to include author information, even for blind
% submissions: the style file will automatically remove it for you
% unless you've provided the [accepted] option to the icml2025
% package.

% List of affiliations: The first argument should be a (short)
% identifier you will use later to specify author affiliations
% Academic affiliations should list Department, University, City, Region, Country
% Industry affiliations should list Company, City, Region, Country

% You can specify symbols, otherwise they are numbered in order.
% Ideally, you should not use this facility. Affiliations will be numbered
% in order of appearance and this is the preferred way.
\icmlsetsymbol{equal}{*}

\begin{icmlauthorlist}
\icmlauthor{Muhammad Mubashar}{yyy}
\icmlauthor{Shireen Kudukkil Manchingal}{yyy}
\icmlauthor{Fabio Cuzzolin}{yyy}
% \icmlauthor{Firstname1 Lastname1}{equal,yyy}
% \icmlauthor{Firstname2 Lastname2}{equal,yyy,comp}

% \icmlauthor{Firstname3 Lastname3}{comp}
% \icmlauthor{Firstname4 Lastname4}{sch}
% \icmlauthor{Firstname5 Lastname5}{yyy}
% \icmlauthor{Firstname6 Lastname6}{sch,yyy,comp}
% \icmlauthor{Firstname7 Lastname7}{comp}
% %\icmlauthor{}{sch}
% \icmlauthor{Firstname8 Lastname8}{sch}
% \icmlauthor{Firstname8 Lastname8}{yyy,comp}
%\icmlauthor{}{sch}
%\icmlauthor{}{sch}
\end{icmlauthorlist}

\icmlaffiliation{yyy}{School of Engineering, Computing and Mathematics, Oxford Brookes University, UK}
% \icmlaffiliation{comp}{Company Name, Location, Country}
% \icmlaffiliation{sch}{School of ZZZ, Institute of WWW, Location, Country}

\icmlcorrespondingauthor{Muhammad Mubashar}{19230664@brookes.ac.uk}
% \icmlcorrespondingauthor{Firstname2 Lastname2}{first2.last2@www.uk}

% You may provide any keywords that you
% find helpful for describing your paper; these are used to populate
% the "keywords" metadata in the PDF but will not be shown in the document
\icmlkeywords{Machine Learning, ICML}

\vskip 0.3in
]

% this must go after the closing bracket ] following \twocolumn[ ...

% This command actually creates the footnote in the first column
% listing the affiliations and the copyright notice.
% The command takes one argument, which is text to display at the start of the footnote.
% The \icmlEqualContribution command is standard text for equal contribution.
% Remove it (just {}) if you do not need this facility.

\printAffiliationsAndNotice{}  % leave blank if no need to mention equal contribution
% \printAffiliationsAndNotice{\icmlEqualContribution} % otherwise use the standard text.
\begin{abstract}
Large Language Models (LLMs) are known to produce very high-quality tests and responses to our queries. But how much can we trust this generated text? In this paper, we study the problem of uncertainty quantification in LLMs. We propose a novel Random-Set Large Language Model (RS-LLM) approach which predicts finite random sets (belief functions) 
%based on Dempster-Shafer theory of evidence, 
over %a set of tokens, 
the token space,
rather than probability vectors 
%over the tokens themselves 
as in classical LLMs. In order to allow so efficiently, we also present a methodology based on hierarchical clustering to extract and use a budget of ``focal" subsets of tokens upon which the belief prediction is defined,
%from the token space 
rather than using all possible collections of tokens, making the method scalable yet effective. RS-LLMs encode the epistemic uncertainty induced in their generation process by the size and diversity of its training set via the size of the credal sets  associated with the predicted belief functions. 
% {\color{cyan} Experiments to be added.}
The proposed approach is evaluated on CoQA and OBQA datasets using Llama2-7b, Mistral-7b and Phi-2 models and is shown to outperform the standard model in both datasets in terms of correctness of answer while also showing potential in estimating the second level uncertainty in its predictions and providing the capability to detect when its hallucinating.
\end{abstract}

\section{Introduction} \label{sec:introduction}

Large Language Models (LLMs) \citep{achiam2023gpt, anil2023palm, touvron2023llama} have made significant waves in the recent years and have shown to perform very well on a wide range of NLP tasks like question-answering \cite{wu2023brief}, common sense reasoning \cite{wei2022chain}, mathematical problem-solving \cite{lewkowycz2022solving} and code generation \cite{roziere2023code}. Typically, these models are pre-trained on a large corpus of text to predict the next token in an unsupervised fashion. To make these models usable, they are further fine-tuned on a particular application \cite{cobbe2021training} and aligned to make sure the model behaves in an ethical manner and according to the human preferences \cite{bai2022training}.

However, LLMs still have limitations in their capacity to understand information and often produce false statements or “hallucinations” \cite{maynez2020faithfulness}. This makes them less trustworthy and causes hindrance to deploying LLMs in high-stake decision-making applications where the consequences of incorrect decisions are severe. Therefore, there needs to be a mechanism to make LLMs more truthful (calibration) and to associate uncertainty estimation with the model’s generation. Furthermore, they need to distinguish the source of this uncertainty \cite{cuzzolin2024uncertainty}, as the latter can be of either aleatoric (relating to chance) or epistemic (relating to knowledge) nature \citep{kendall2017uncertainties, hullermeier2021aleatoric}. However, LLMs, as they are now, lack the ability to do so.

Next-token prediction in language models can be thought of simply as classification over the all the tokens in the vocabulary. Therefore, it appears natural to apply to language models uncertainty quantification approaches developed for classification tasks, while paying attention to scalability. 
\\
Motivated by the introduction of Random-set Neural Networks (RS-NN) \cite{RS-NN} in classification, which demonstrated superior performance compared to Bayesian, ensemble and traditional approach in terms of accuracy, robustness and uncertainty quantification, here we propose Random-set Large Language Models (RS-LLMs). Random sets have been proposed as a mathematical model for subjective belief, alternative to Bayesian probability \citep{Molchanov05,Molchanov_2017, cuzzolin2023reasoning}. 
%Random sets 
They 
assign probabilities to sets of classes directly, in a non-additive way, and can therefore naturally model the fact that observations come in the form of sets, %especially in the case of confusion. 
e.g. when data components are missing either in the feature space or in time.
As the vocabulary in LLMs lives in a finite space, we model the random sets using \emph{belief functions} \cite{Shafer76}, the finite incarnation of random sets \cite{cuzzolin2018visions,cuzzolin2020geometry,cuzzolin2001geometric,cuzzolin2010geometry}.

\begin{figure*}[t!]
\centering
\includegraphics[width = 0.9\textwidth]{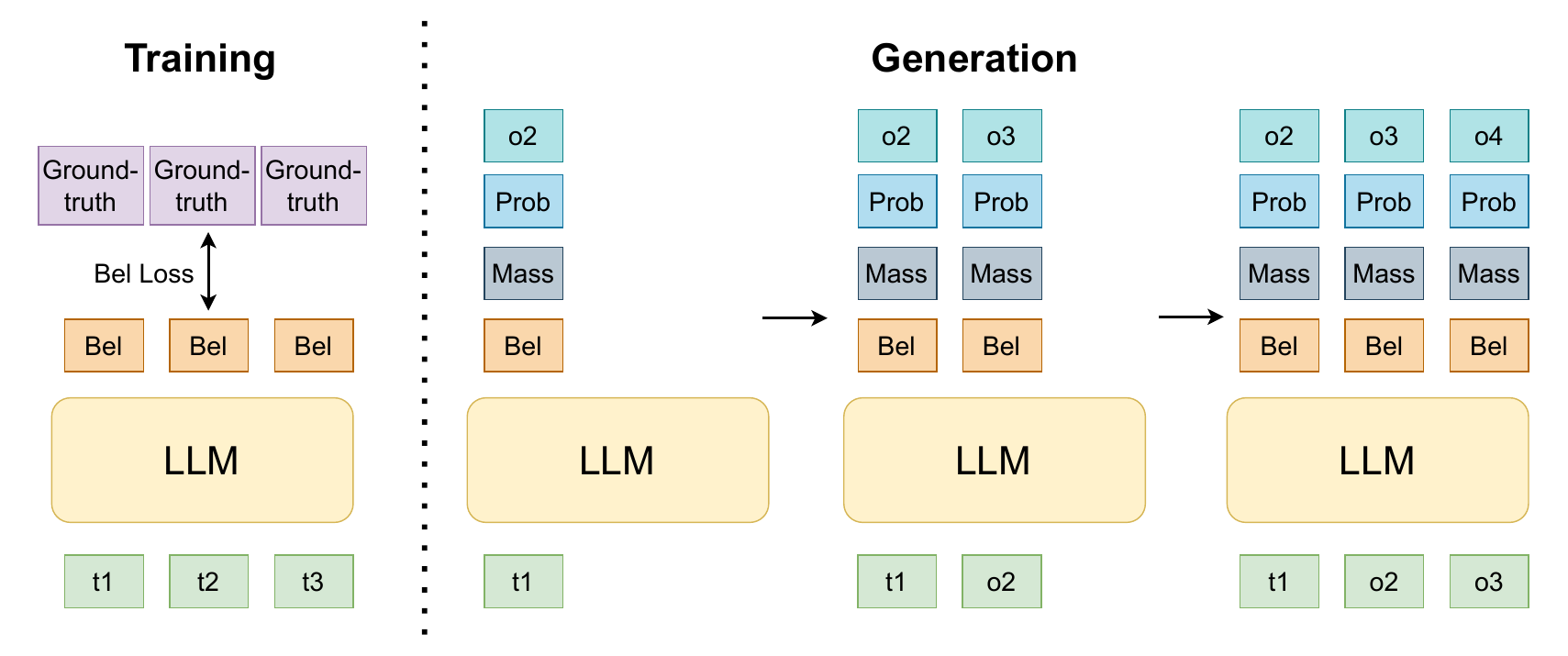}
\caption{Training and generation flow of RS-LLM. Training is performed in a parallel fashion using the teacher forcing method. Generation is done sequentially. For each token, the model predicts a belief function. Then the mass function, probability distribution and next token is subsequently computed/sampled from that belief function.
% {\color{red}I moved the diagram to the Introduction, for people expect this nowadays. Makes sure you use a consistent face and size of font across all figures.}
} \label{fig:rsllm-flow}
\end{figure*}

Instead of predicting a probability distribution over the vocabulary for the next token, an RS-LLM predicts a belief function \cite{cuzzolin2010three} over the vocabulary. This belief function assigns a belief value $bel(A) \in [0,1]$ to all $A \in \mathcal{O}$, where $\mathcal{O}$ is a finite budget of sets of tokens in the vocabulary. This finite budget of focal sets is selected by hierarchical clustering \citep{mullner2011modern, ackermann2014analysis}
% {\color{red}add ref}
over the embedding of all the tokens. Such a predicted belief function \cite{cuzzolin2014belief} is mathematically equivalent to a convex set of probability vectors (\emph{credal set} \citep{levi80book,cuzzolin2008credal}) on the vocabulary of tokens and can be interchangeably be represented by a mass function with  mass values $m(A) \in [0,1] \; \forall \; A \in \mathcal{O}$. The center of mass of this mass function gives a probability distribution, termed as \emph{pignistic probability} \citep{smets1994transferable}. The uncertainty associated with a particular output can then be expressed by either the entropy of the pignistic prediction or by the width of the credal prediction. The uncertainty associated with the whole generated sentence can then be computed as the maximum of the uncertainties among all predicted tokens.

The benefit of having a belief function as output, as a way of expressing epistemic uncertainty about the prediction, can be illustrated using an example. Suppose that a standard LLM was tasked to predict the next token in the sentence \textit{Joe likes to play \_\_\_} given the vocabulary $\mathcal{V} = \{baseball, basketball, \cdots \}$ and it predicted $\mathcal{P}:$ $\{ P(baseball) = 0.5, P(basketball) = 0.5 \}$. So, there is a 50\% chance of the next token being baseball and 50\% of being basketball. However, this does not tell us if this predicted probability distribution is equally distributed because the LLM did not quite know the answer, or if both answers are valid.
%In opposition, if the model were to predict belief functions, the same probability distribution can be represented by a multitude of belief functions each having a different interpretation. 
Allowing the model to predict a belief function, in opposition, gives the LLM more expressive power to clarify the source of the uncertainty.

Consider the two belief functions 

$Bel\_1: \{bel(\{baseball\} = 0, \; bel(\{basketball\} = 0, \; bel(\{baseball, basketball\} = 1 \}$ and 

$Bel\_2: \{bel(\{baseball\}=0.5, \; bel(\{basketball\} = 0.5, \; bel(\{baseball, basketball\} = 1 \}$. 

As it can easily be shown, both 
%of these belief functions 
have the above predicted probability distribution $\mathcal{P}$ as their center of mass (pignistic distribution). So, 
%so the end result will be same, 
the corresponding pignistic prediction would be the same. However, we can now associate an explanation with that answer. If the model predicts $Bel\_1$, this means that the LLM does not know which one of baseball or basketball Joe likes, only that the answer is in the set ${baseball, basketball}$. Whereas, predicting $Bel\_2$ means that Joe likes both baseball and basketball equally, so either can be the answer. 
%This example clearly demonstrates the benefit of predicting belief functions instead of probability distributions.

% {\color{red}Briefly outline how training and generation work in RS-LLM, referring to Fig. 1.}

% {\color{red}Based on what you say in the experiments, we need to highlight our proposal is unique among UQ approaches to LLM!}
% {\color{cyan} Will do.}

% {\color{cyan}
Fig. \ref{fig:rsllm-flow} encapsulates the training and generation process of RS-LLM. At training time, the RS-LLM is tasked to predict a belief function and is updated using the loss function highlighted in Sec \ref{sec:loss}. At generation time, the RS-LLM once again predicts a belief function, which is then transformed into a mass function. Its pignistic probability is then computed to obtain the final token. The training is done in parallel fashion using the teacher forcing method \citep{williams1989learning} while generation needs to be done sequentially token by token. Our method is the first and only one that, as shown by our experiments, uniquely provides the LLMs with the capability to deal with second level uncertainty, so boosting its accuracy, while also providing tools to reason about hallucination by using Pignistic entropy and Credal Set Width Sec. \ref{uncertainty-estimation}.
% }

\textbf{Contributions}. This paper makes the following contributions to the wider research in LLMs: % through this paper:
\begin{enumerate}
    \item We propose a unique Random-set Large Language Model (RS-LLM) approach based on predicting belief functions over the vocabulary instead of just a probability distribution. This leads to better uncertainty estimation and explainability of an LLM's generation.
    \item We advance an intuitive approach for selecting the budget of the focal sets of hierarchical clustering over the embedding of the tokens, mitigating the computational complexity associated with working with random sets and making the approach scalable.
    \item We set out a method to measure the uncertainty associated with an LLM's generation process using both the entropy of pignistic prediction and the width of the credal set \cite{wang2024credal,caprio2024credal,wang2025creinns,wang2024credal} associated with the predicted belief function.
    \item We show significant experimental results of our approach while also exploring it in the context of hallucination. RS-LLMs outperforms Standard LLMs in terms of closeness to the groundtruth, while also displaying remarkable capabilities in hallucination detection using credal set width.
    % {\color{red}I suppose this is still to be finished? You need to detail what experiments, and in what way RS-LLM is superior. What do they demonstrate?} {\color{cyan} Done}
\end{enumerate}

\textbf{Paper Outline.} This paper is structured 
%in the following manner. 
as follows.
Sec. \ref{sec:introduction} introduces the need for and concept of Random-Set Large Language Model. Sec. \ref{sec:related_work} outlines the existing approaches and methods for uncertainty quantification in LLMs. In Sec. \ref{sec:background}, we recall the notions of random sets, belief functions, credal and pignistic
predictions. Sec. \ref{sec:rsllm} details our RS-LLM approach, loss and uncertainty representation. Finally, Sec. \ref{sec:experiments} provides empirical evidence supporting out approach and discusses the results. Conclusions are given in Sec. \ref{sec:conclusions}.

\section{Related Work} \label{sec:related_work}
Uncertainty quantification in LLMs is a hot topic and indeed, significant work has recently been done in this regard. \cite{plaut2024softmax} proposes to simply use softmax entropy quantification, whereas \citep{kuhn2023semantic, farquhar2024detecting} introduce a semantic entropy, measured over multiple runs of the model, as the uncertainty measure. Some methods use calibration for uncertainty evaluation \citep{glushkova2021uncertainty, wang2022uncertainty} and use calibration errors to evaluate if the models can be trusted. One stream of methods \cite{chen2024inside} uses the hidden state information within the models to quantify uncertainty. Other methods address uncertainty by prompting the model to evaluate its own generation \cite{mielke2020linguistic} or by fine-tuning the model so that it predicts its own uncertainty \citep{lin2022teaching, kadavath2022language}. In addition, general-purpose methods like Monte-Carlo Dropout (MCD) \cite{gal2016dropout}, Bayes by Backprop (BBB) \cite{blundell2015weight} and Deep Ensembles \cite{balabanov2024uncertainty} have also been applied to LLMs to quantify uncertainty. Laplace-LORA \cite{yang2023bayesian} and BLoB \cite{wang2024blob} apply Bayesian methods to LoRA \cite{hu2021lora}. LoRA provides an efficient way of fine-tuning LLMs by injecting low-rank weight matrices into transfomer layers and training them instead of the whole network.
% {\color{red}Briefly explain LoRA}. 
Finally, ENN-LLM \cite{osband2022fine} measures uncertainty by using their ensemble-inspired Epinet with LLMs.

% {\color{cyan}
However, we do not compare ourselves with these uncertainty methods as they (i) mostly measure uncertainty by modifying the prompt or over multiple runs \citep{mielke2020linguistic, osband2022fine, kuhn2023semantic, farquhar2024detecting, wang2024blob, farquhar2024detecting} and (ii) because our method is a complete training regimen rather than just being a wrapper method \citep{mielke2020linguistic, glushkova2021uncertainty, wang2022uncertainty, plaut2024softmax}. \citep{lin2022teaching, kadavath2022language} and \cite{chen2024inside} directly train their model to predict uncertainty or use hidden state information to compute uncertainty, which is not directly comparable to our method.
% }

% {\color{red}Feels a bit short.. having space you could beef it up. Also, you do need to contrast our approach to those of these competitors.}
% {\color{cyan} Done}
% Laplace-LORA \cite{yang2023bayesian} and BLoB \cite{wang2024blob} employ Bayesian methods over LoRA \cite{hu2021lora}. Similarly, methods like Monte-Carlo Dropout (MCD) \cite{gal2016dropout}, Bayes by Backprop (BBB) \cite{blundell2015weight} and Deep Ensembles \cite{balabanov2024uncertainty} have also been used over LLMs to quantify uncertainty. ENN-LLM \cite{osband2022fine} measures uncertainty by using their ensemble inspired Epinet with LLMs. One stream of methods \cite{chen2024inside} use hidden state information of models to quantify uncertainty while others \cite{plaut2024softmax} just leverage the softmax entropy of the model output. However, almost all of these methods either suffer in inference times (due to the need running multiple times) or in terms of performance.

\section{Random sets and belief functions}\label{sec:background}

\textbf{Random sets.} Consider a sensor measuring a 2D position vector, $\mathbf{x} = (x, y)$, where each component represents a spatial coordinate, modelled by a data probability distribution. Ideally, the sensor outputs precise values for each coordinate. However, suppose due to occlusion or sensor failure, the second coordinate $y$ is unreadable. 

Instead of assigning a single value, we can model the missing component $y$ as the set $Y$ containing all possible values for that component. The underlying random process that generates the data remains, but the resulting random variable assumes values which can be sets:
\begin{equation}
\mathbf{X} = (x, Y),
\end{equation}
where $Y$ is a set accounting for the missing information. % (see Fig. ).
%within a given range $[a, b]$. 
%The observed vector is then transformed into a \emph{random set} representation:
The data generation process can thus be modelled as a set-valued random variable, i.e., a \emph{random set}.
Random sets generalise classical additive probability measures by representing uncertainty through sets, rather than precise values, making random sets a powerful tool for modeling imprecise observations \citep{Matheron75, Nguyen78, Molchanov05}. 

% A die is a simple example
% of a (discrete) random variable. Its probability space is defined on the sample space $\Theta = \{ \text{face1}, \text{face 2}, \ldots , \text{face 6}\}$, where elements are mapped to the real numbers $1,2,\ldots,6$, respectively.
% Now, imagine that faces 1 and 2 are cloaked, and we roll the die. How do we model this new experiment, mathematically? Actually, the probability space has not changed (as the physical die has not been altered, its faces still have the same probabilities). What has changed is the mapping: since we cannot observe the outcome when a cloaked face is shown 
% (assuming that only the top face is observable),
% both face 1 and face 2 (as elements of $\Theta$) are mapped to the set of possible values $\{1,2\}$ on the real line $\mathbb{R}$ (Fig. \ref{fig:cloaked-die}). Mathematically, this is a \emph{random set} \citep{Matheron75,Kendall74foundations,Nguyen78,Molchanov05}, i.e., a set-valued
% random variable, modelling random experiments in which observations come in the form of sets.

\iffalse
\begin{figure}[h]
    \centering
    \includegraphics[width =0.9\linewidth]{figures/RS_LLM_Random.pdf} 
    \caption{The random set associated with occlusion on sensor reading. {\color{red}I am not sure they need this to understand - use the space to provide more results or details, rather.}} \label{fig:sensor_occlusion}
\end{figure}
\fi

\begin{figure}[h]
    % \centering
    \hspace{-23pt}
    \includegraphics[width = 1.14\linewidth]{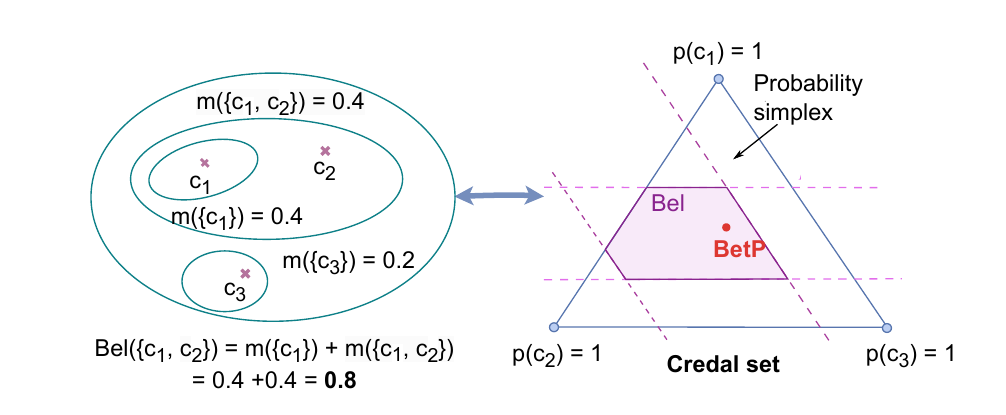}
    \vspace{-6pt}
    \caption{A belief function measures the total belief (sum of masses of its subsets) for a set (Eq. \ref{eq:belief-mobius}). 
    \label{fig:belief-credal} 
    % \textcolor{orange}{Changed figure.}
    % {\color{red}What do the (overlapping) circles represent? Not the focal sets. The older picture was better in that sense, please use a similar style. Also remove "Belief Values"}
    }
\end{figure}

\textbf{Belief functions.} 
Random sets have been proposed by \citet{dempster2008upper} and \citet{Shafer76} as a mathematical model for subjective belief, alternative to Bayesian reasoning. 
Thus,
on finite domains (e.g., for classification) they assume the name of \emph{belief functions}.
While classical discrete mass functions assign normalised, non-negative values to \emph{elements} $\theta \in \Theta$ of their sample space, a belief function independently assigns normalised, non-negative mass values to its \emph{subsets} $A \subset \Theta$: 
\begin{equation}\label{eq:mass-non-neg}
m(A) \geq 0, \: \forall A \subset \Theta, \quad
\sum_A m(A)=1. %\vspace{-6pt}
\end{equation} 
The {belief function} %(BF) 
associated with a mass function $m$
measures the total mass of the subsets of each \emph{focal set} $A$. Mass functions can be recovered from belief functions via Moebius inversion \citep{Shafer76}, 
which, in combinatorics, plays a role similar to that of the derivative in calculus:
\begin{equation} \label{eq:belief-mobius}
Bel(A) = \sum_{B\subseteq A} m(B), \:\:\:\:\:
m(A) = \sum_{B\subseteq A}{(-1)^{|A\setminus B|} {Bel}(B)}.
\vspace{-6pt}
\end{equation}

For example, consider a classification problem where an object belongs to one of three possible categories, $\Theta = \{c_1, c_2, c_3\}$. A belief function might express uncertainty by assigning mass as follows:
\begin{equation}
m(\{c_1\}) = 0.4, m(\{c_3\}) = 0.2, m(\{c_1, c_2\}) = 0.4.
\end{equation}
Here, 40\% of the belief supports $c_1$, 20\% supports $c_3$, and 40\% supports the composite hypothesis that the object belongs to either $c_1$ or $c_2$ but not $c_3$, without being able to specify which. The belief value $Bel(A)$ of a set of classes $A$ accumulates mass from all the subsets of $A$:
\begin{equation}
Bel(\{c_1, c_2\}) = m(\{c_1\}) + m(\{c_1, c_2\}) = 0.4 + 0.4 = 0.8.
\end{equation}
This means we have an 80\% belief that the object belongs to either $c_1$ or $c_2$, reflecting epistemic uncertainty rather than simply assigning a probability to each class. 
Belief functions generalise Bayesian probability by allowing explicit representation of uncertainty, making them useful in applications where knowledge is incomplete or ambiguous \cite{shafer1976a, bouckaert1995bayesian}. 
% {\color{red}I think we could add some references in support here}

Belief functions have been employed in the past for machine learning purposes, e.g. classification \cite{cuzzolin2018generalised} regression \cite{gong2017belief,cuzzolin2000integrating}.

\textbf{Credal prediction.} As anticipated, RS-LLM is designed to predict a belief function (a finite random set) on the set of classes. 
A belief function, in turn, is associated with a convex set of probability distributions (a \emph{credal set} \citep{levi80book,zaffalon-treebased,cuzzolin2010credal,antonucci2010credal,cuzzolin2008credal}) on the same domain. 
This is the set:
\begin{equation} \label{eq:consistent}
Cre
=  \left \{ P : 
\Theta
\rightarrow [0,1]
| Bel(A) \leq P(A) 
%\leq Pl(A) 
\right \},
%{-5pt}
\end{equation}
of probability distributions $P$ on $\Theta$ which dominate the belief function on each and every focal set $A$.

The size of the resulting credal prediction measures the extent of the related epistemic uncertainty (see Sec. \ref{uncertainty-estimation}). Indeed, the use of credal set size as a measure of epistemic uncertainty is well-supported in literature \citep{hullermeier2021aleatoric,bronevich2008axioms}, as it aligns with established concepts of uncertainty such as conflict and non-specificity \citep{yager2008entropy,kolmogorov1965three}.
%A credal prediction, by encompassing multiple potential distributions, reflects the model's acknowledgment of this uncertainty. 
A wider credal set indicates higher uncertainty, as the model refrains from committing to a specific probability distribution due to limited or conflicting evidence. In contrast, a narrower credal set implies lower uncertainty, signifying a more confident prediction based on substantial, consistent evidence.

\textbf{Pignistic probability}. The 
\emph{pignistic probability} is the probability distribution obtained by re-distributing the masses of all the focal sets $A$ of a belief function $Bel$ to their constituent elements 
%obtained from a belief function, 
$\theta \in A$:
\begin{equation}\label{eq:pignistic}
    BetP(\theta) = \sum_{A \ni \theta} \frac{m(A)}{|A|}.
\end{equation}
Smets \citep{smets2005decision} originally proposed to use the pignistic probability for decision making using belief functions, by applying expected utility to it. Notably, the pignistic probability is geometrically the centre of mass of the credal set 
(\ref{eq:consistent})
%(see Fig. \ref{fig:belief-credal}) {\color{red}The illustration of credal set is not there anymore; do we need it?} 
associated with a belief function \citep{cuzzolin2008credal,cuzzolin2018visions,cuzzolin2020geometry}. 

\section{Random-Set Large Language Models }\label{sec:rsllm}

\subsection{Representation} \label{sec:arch}

RS-LLM predicts for each input sequence of tokens a 
%mass/
belief function for the next token,
%set-valued observations of each sample in the dataset,
rather than a vector of softmax probabilities as in a traditional LLM. 
%Namely, it outputs a score for a collection of sets of classes
A `vanilla' RS-LLM, having a vocabulary of $T$ tokens, would have $2^N$ outputs (as $2^N$ is the cardinality of $\mathbb{P}(T)$), each being the belief value of the focal set of tokens $A \in\mathbb{P}(T)$ corresponding to that output neuron. 
This would requires the output layer of the LLM to be amended to have $2^N$ outputs, as the only change needed to convert any LLM architecture to an RS-LLM architecture.
In practice, for efficiency only a budget of focal sets is identified from the training data (Sec. \ref{sec:Budgetting}).

Since the model is tasked to predict a belief function, the ground-truth for training needs to be modified. Given an input sequence, the ground-truth in a classical LLM is the one-hot encoding of the next token. The belief-encoded ground-truth for RS-LLM, instead, is given by the vector
$\mathbf{bel} = \{ Bel(A), A \in \mathbb{P}(T) \}$
of belief values for each focal set of tokens $A \in \mathbb{P}(T)$.
$Bel(A)$ is set to 1 iff the true next token is contained in the subset $A$, 0 otherwise.
This corresponds to full certainty that the element belongs to that set and there is complete confidence in this proposition.
Fig. \ref{fig:rsllm-flow} shows the training and generation flow of RS-LLM 
% \textcolor{cyan}{
(detailed in Sec. \ref{sec:introduction})
% }. 
% {\color{red}Are these also described in the text somewhere? See my comment in the Introduction. Should we mention that ours is a fine-tuning approach and mention LoRA?}
% {\color{magenta} I have describe it in the introduction and also refered it here}
% {\color{magenta} I think we shouldn't put too much emphasis on LoRA. As it is just a mechanism we employ to add more weights to the network training. I do mention that we use it in the experiment section.}

\subsection{Budgeting} \label{sec:Budgetting}

Random-sets are defined on the power set of all subsets of their domain. 
%work over all the subset domain but 
However, this is computationally impossible 
%for cases when there are large number of elements, 
when the domain cardinality is extremely large, as is the case 
%especially 
for large language models where the typical size of vocabulary is $\sim$32K tokens. 
The number of subsets in this case would be an astronomical $2^{32K}$. % which is an astronomically large number. 
Therefore, to tackle this problem, we need a method to select a budget of most appropriate focal sets and use only those as our focal sets. 

Inspired by \cite{RS-NN,manchingal2023random}, we propose a strategy based on clustering the most similar tokens. This is appropriate because, by intuition, the model will most likely be confused among tokens which are similar to each other 
% {\color{cyan}
and potentially have similar semantics \cite{farquhar2024detecting}.
%{\color{magenta} (and potentially have a similar semantics, ref Gal Nature paper?)}. 

As a first step, token embeddings are computed using a pre-trained LLM (in fact, any method for obtaining embedding of tokens can be used here). Then, hierarchical clustering is applied to cluster similar tokens. The number of clusters $K$ is a hyper-parameter which determines the granularity of the belief function.

\begin{figure*}[ht]
\centering
\includegraphics[width = 0.9\textwidth]{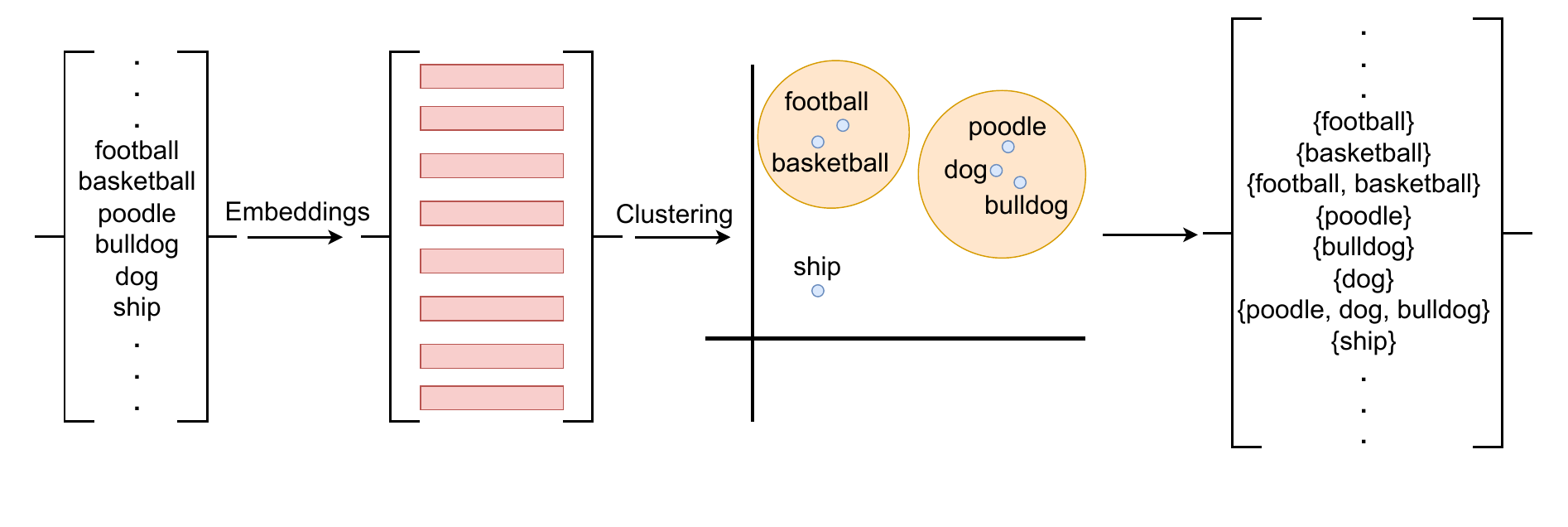}
\vspace{-8mm}
\caption{Proposed budgeting method for RS-LLM. First, embeddings are computed for all the tokens in vocabulary. Then, focal sets are computed using hierarchical clustering.} \label{fig:rsllm-budgeting}
\end{figure*}

The final budget $\mathcal{O}$ of focal sets of the belief functions to be predicted is the union of the clusters of tokens so obtained and the collection of singleton sets containing the $T$ original tokens, amounting to a total of $(K + T)$ sets.  Fig. \ref{fig:rsllm-budgeting} shows a detailed overview of the proposed budgeting method. An analysis of the focal sets obtained by this budgeting strategy is presented in Appendix \ref{sec:budgeting-analysis}.

\subsection{Loss function} 
\label{sec:loss} %Loss function

% {\color{red}I am a bit conscious that the loss function is exactly the same as in RS-NN, which might mark us down re novelty.}
% {\color{magenta} Yes the only difference is that we don't use moebius inverse. the rest is almost same. I actually tried playing around with the loss like making BCE itself weighted or changing the formulation of $M_s$ term. But eventually found that the current is the best} 
As highlighted in \cite{RS-NN}, a random-set prediction problem is mathematically similar to the multi-label classification problem, for in both cases the corresponds to a vector of 0s and 1s and the prediction is vector in $[0,1]$. They do have different semantics, as in the case of random sets, they indicate if the set contains true class or not; whereas in multi-label classification, it is the probability of input belonging to a particular class.

Despite the different semantics, we can thus adopt as loss Binary Cross-Entropy (BCE) (\ref{bce-loss}) 
with sigmoid activation, 
to drive the prediction of a belief value
for each focal set in the identified budget:
% {\color{red}This equation looks pretty ugly}:
% {\color{cyan} Yesh, the problem is that it's very long. I will try to see if I can make it better} DONE

\vspace{-10pt}
\begin{equation}\label{bce-loss}
\hspace{-5pt}
\begin{array}{lll}
    \mathcal{L}_{BCE} = \displaystyle
    - \frac{1}{b_{size}} \sum_{i=1}^{b_{size}} 
    \frac{1}{seq\_len} \sum_{j=1}^{seq\_len} 
    \frac{1}{|\mathcal{O}|}
    \sum_{A \in \mathcal{O}}
    \displaystyle 
    \Big [
    Bel_{ij}(A) 
    \\ \vspace{-6pt} \\
    \quad \log(\hat{Bel}_{ij}(A))
   \displaystyle 
   + (1 - Bel_{ij}(A)) 
   \log(1 - \hat{Bel}_{ij}(A)) \Big ].
\end{array}
\end{equation}
Here, $i$ is the index of the training text and $j$ is the index of the token in that text of length $seq\_len$ within a batch of cardinality $b_{size}$.
$A$ is a focal set of classes in the budget $\mathcal{O}$.
$Bel_{ij}(A)$ is the $A$-th component of the vector $\mathbf{bel}_{ij}$ encoding the ground truth belief values
for the $j$-th token of $i$-th training text,
and $\hat{Bel}_{ij}(A)$ is the corresponding belief value in the predicted vector $\hat{\mathbf{bel}}_{ij}$ for the same training point. 
Both $\mathbf{bel}_{ij}$ and $\hat{\mathbf{bel}}_{ij}$ are vectors of cardinality $|\mathcal{O}|$ for all $i$ and $j$.
%{-3pt}

% \textcolor{blue}{
To be valid, a belief function needs to be such that the corresponding mass values are non-negative and should sum up to 1 \citep{Shafer76} (see Eq. (\ref{eq:mass-non-neg})).
%A valid belief function comes with further conditions and satisfies Eq. \ref{eq:mass-non-neg} which states that mass values derived from belief functions should be non-negative and should sum up to 1 \citep{Shafer76}. 
To ensure that the predicted scores amount to a valid belief function,
a mass regularisation term $M_r$ (\ref{eq:mass-reg}) and a mass-sum term $M_s$ (\ref{eq:mass-sum}) are incorporated into the loss function. These terms encourages non-negativity and normalization (sum equal to 1) of the (predicted) mass values $\hat{m}(A)$, $A \in \mathcal{O}$ respectively:\footnote{At any rate, improper belief functions are normally used in the literature \citep{denoeux2021distributed}, e.g. for conditioning \citep{cuzzolin2010geometric,cuzzolin2011geometric}.}
\begin{equation} \label{eq:mass-reg}
\begin{array}{l}
\displaystyle
    M_s = \max \left(0, \frac{1}{b_{size}} \sum_{i=1}^{b_{size}}\sum_{A \in \mathcal{O}} \hat{m}_i(A) - 1 \right), \hspace{0.5em}
\end{array}
\end{equation}
\begin{equation} \label{eq:mass-sum}
\begin{array}{l}
\displaystyle
    M_r = \frac{1}{b_{size}} \sum_{i=1}^{b_{size}}\sum_{A \in \mathcal{O}} \max(0, - \hat{m}_i(A)).
\end{array}
\end{equation}
These mass values are obtained from the predicted belief function $\hat{\mathbf{bel}}$ via the rearranging of the belief definition, i.e.: %as in Eq. (\ref{eq:bel-to-mass}):
\begin{equation} \label{eq:bel-to-mass}
\begin{array}{l}
\displaystyle
    m(A) = Bel(A) - \sum_{B \subset A} m(A).
\end{array}
\end{equation}
The reason is that the Moebius inverse cannot be used here, as it requires the belief values for all the subsets of a given set to compute the corresponding mass function.

All loss components $\mathcal{L}_{BCE}$, $M_r$ and $M_s$ are computed during batch training. 
% {\color{red}justification?} 
Two hyperparameters, $\alpha$ and $\beta$, control the relative importance of the two mass terms, %within % are added to the BCE loss $\mathcal{L}_{BCE}$, 
yielding as the total loss for RS-LLM:
\begin{equation}\label{final_loss}
    \mathcal{L}_{RS} = \mathcal{L}_{BCE} + \alpha M_r + \beta M_s.
\end{equation}

To ensure adherence to valid belief functions, regularisation terms are introduced to discourage deviations, akin to how training-time regularisation in neurosymbolic learning fosters predictions that align with commonsense reasoning \citep{giunchiglia2023road}. 
% Prior studies indicate that imposing soft constraints \citep{marquez2017imposing} can be just as effective as enforcing hard constraints.
Nevertheless, when $\alpha$ and $\beta$ are too small, regularisation alone may not suffice to maintain the validity of belief-function predictions. In such instances, a corrective post-processing step is implemented: negative mass values are reset to zero, and an additional `universal' set — comprising all classes — is incorporated into the final budget. This ensures that any residual mass is assigned to this set, preserving the requirement that the total mass across all focal sets in $\mathcal{O}$ sums to 1. This adjustment follows well-established approximation methods \cite{cuzzolin2009complexes,cuzzolin2011consistent}, as detailed in \citet{cuzzolin2020geometry}, Part III or \cite{cuzzolin2013l_}.
% }

\subsection{Uncertainty estimation}\label{uncertainty-estimation}
\textbf{Pignistic prediction}.
% The pignistic probability (Eq. \ref{eq:pignistic}) is the central prediction associated with a belief function (seen as a credal set).
% Even though this behaves like a standard probibility distribution, the RS-LLM architecture and training mechanism are designed to facilitate set-based learning by incorporating class set information during training. Hence, when pignistic probabilities are computed, they reflect the learning derived from the masses and beliefs of various subsets, making it more reliable than traditional softmax probabilities (as our experiments show).
The pignistic probability (Eq. \ref{eq:pignistic}) \cite{SMETS2005133} provides a representative estimate derived from a belief function, which can be understood as defining a credal set. Although it resembles a conventional probability distribution in behavior, the RS-LLM framework is explicitly designed to incorporate class set relationships during training, fostering a more structured form of learning. As a result, the pignistic probabilities produced by this model encode information from the underlying mass functions and subset structures, leading to predictions that, as shown in our experiments, offer greater reliability than those obtained through traditional softmax outputs.

\textbf{Entropy of the pignistic prediction}.  
%Entropy for RS-NN is computed using 
The Shannon entropy of the pignistic prediction $BetP_j$ for a particular token at index $j$ can then be used as a measure of the uncertainty associated with that token. This is in some way analogous to the entropy of a softmax probability prediction.

The cumulative uncertainty associated with the whole generated text is the mean uncertainty of that text.
\begin{equation}\label{entropy-pred}
    H_{RS} = mean \left \{-\sum_{t \in T}{BetP_j}(t)\log {BetP_j}(t) \right \}.
\end{equation} 
A higher entropy value (\ref{entropy-pred}) thus indicates greater uncertainty in the model's predictions.

\textbf{Size of the credal prediction}. A meaningful way to quantify the epistemic uncertainty in a random-set prediction $\mathbf{\hat{bel}}$ is by assessing the size of its corresponding credal set. Since a credal set forms a convex polytope, various metrics can be used to evaluate its size \citep{sale2023volume}.
For a given token $t$, the probability range within the predicted credal set $\hat{Cre}$ is determined by its upper and lower probability bounds:

\vspace{-5pt}
\begin{equation} \label{eq:size-credal} \overline{P}(t) = \max_{P \in \hat{Cre}} P(t), \quad \underline{P}(t) = \min_{P \in \hat{Cre}} P(t), \vspace{-5pt} \end{equation}

The difference between these bounds, $\overline{P}(t) - \underline{P}(t)$, serves as a measure of the size of the credal set, particularly when considering the most probable next token as determined by the pignistic prediction.
Crucially, for every token $t$, not just the most likely next token $\hat{t}$, the computed pignistic probability estimate $BetP(t)$ is guaranteed to fall within the range $[\underline{P}(t), \overline{P}(t)]$. The width of this interval reflects the degree of epistemic uncertainty in the prediction, with larger intervals indicating greater uncertainty.

\begin{figure*}[ht!]
    \centering
    \includegraphics[width = 0.8\linewidth]{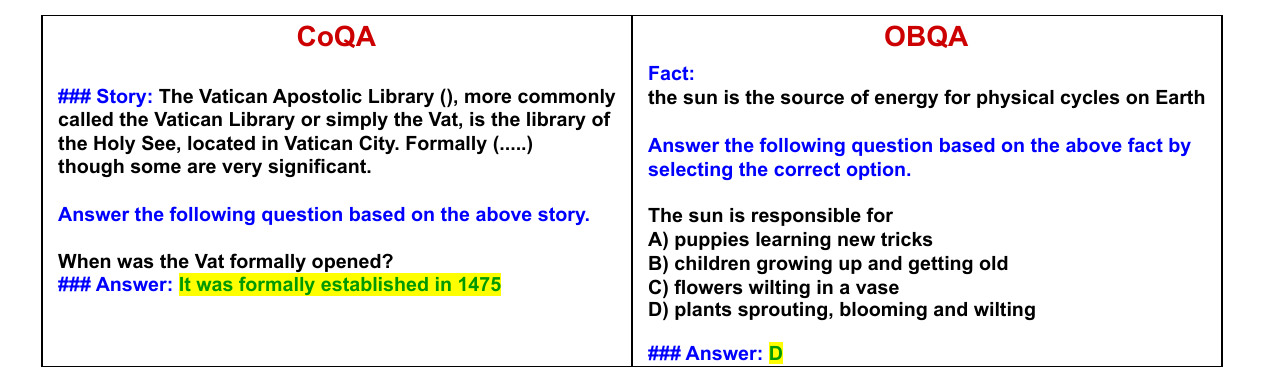}
    \vspace{-1mm}
    \caption{Training examples from CoQA and OBQA datasets. The text in black highlights the actual question, while the blue text represents prompt instructions. The model is trained to predict the text in green.}
    \label{fig:training_samples}
\end{figure*}

\section{Experiments} \label{sec:experiments}
\subsection{Implementation}
\textbf{Datasets.} We evaluate the efficacy of our methods on two Question-Answer Datasets: CoQA \citep{reddy-etal-2019-coqa} and OBQA \citep{OpenBookQA2018}. CoQA is a generative question-answer dataset with the goal of answering a question based on a provided story. OBQA, on the other hand, is a Multiple Choice Question where the aim it select the correct option from a list of options. In this paper, we make use of the OBQA-Additional variant (referred to as OBQA from here on after) of the dataset, which also provides a fact (helpful information) to help answer the question. CoQA consists of 7,200 training samples along with 500 validation samples from 5 different subjects. We use the validation set as the test set. OBQA comprises 4,957 training samples and 500 test samples over STEM subjects.

\textbf{Training Details.} We make use of Llama2-7B-hf \citep{touvron2023llama}, Mistral-7b \citep{jiang2023mistral7b} and Phi-2 \citep{javaheripi2023phi} as our base models. As highlighted in Sec. \ref{sec:arch}, any LLM can be converted to a random-set LLM by simply redefining the last layer to represent sets of tokens instead of individual tokens. This last layer is then trained using the loss function detailed in Sec. \ref{sec:loss}. 

% All experiments are performed on standard Llama2 and Random-Set Llama2 (RS-Llama2). 
%{\color{cyan}
We do not compare ourselves with other uncertainty methods due to the reasons detailed in Sec. \ref{sec:related_work}.
% as they (i) mostly measure uncertainty by modifying the prompt or over multiple runs \citep{mielke2020linguistic,farquhar2024detecting, hu2021lora} and (ii) our method is a complete training regimen rather than just being a wrapper method.
% {\color{red}Does this really hold for all our competitors? You need to clearly explain this in Sec 2 too.}
Both models are trained on NVIDIA A100 80GB GPUs using Huggingface's trl framework \citep{havrilla2023trlx} with default training parameters for 5 epochs, with a batch size of 8. To further boost the training, we add LoRA adapters \citep{yu2023low} of rank 64 to all blocks of the model. Furthermore, we load and train the model in 4-bit mode for enhanced efficiency. 

As anticipated, Llama2-7b, Mistral-7b and Phi-2 serve as the base architectures. By Utilising the budgeting techinique highlighted in Sec. \ref{sec:Budgetting}, we extract $K = 8,000$ focal sets from 32,000 tokens in Llama2 and Mistral, and 51,200 tokens in Phi-2, so that the output size of the last layer for RS-Llama2 and RS-Mistral is $32,000 + 8,000 = 40,000$, and $51,200 + 8,000 = 59,200$ for Phi-2. For RS-Llama2, we set $\alpha = \beta=1e-2$ as hyperparameter values in the loss function.
% {\color{red}Are we doing any ablation on those?}
% {\color{cyan} trying to}

The models are trained using the Supervised Fine-tuning (SFT) method for LLMs. Fig. \ref{fig:training_samples} highlights the training prompt template. The blue text represents the instructions to the LLM while black represents the actual question. The model is trained to predict the text in green. At generation time, the model is given input in the same template but without the answer. The model then continues the statement and produces the answer.

\begin{figure*}[t!]
    \includegraphics[width = \linewidth]{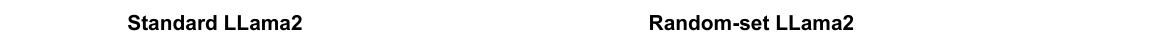}
    \centering
        \begin{subfigure}[t]{0.32\textwidth}
        \centering
        \includegraphics[width = \linewidth]{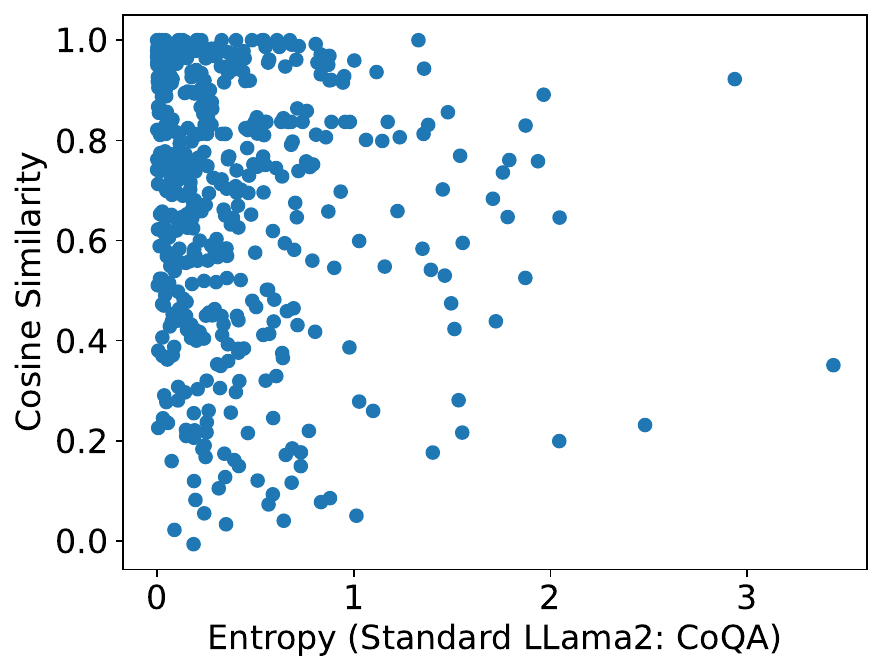}
        \caption{}
        \label{fig:Entropy_CosSim_Standard_LLama2_CoQA}
    \end{subfigure}%
    \begin{subfigure}[t]{0.32\textwidth}
        \centering
        \includegraphics[width = \linewidth]{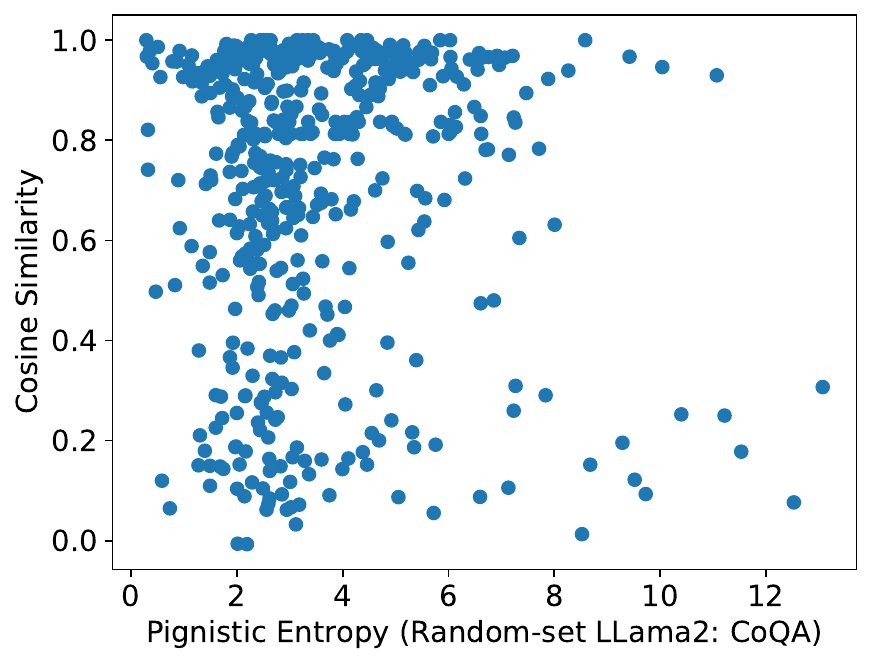}
        \caption{}
        \label{fig:Pig_Entropy_CosSim_RS_LLama2_CoQA}
    \end{subfigure}%       
        \begin{subfigure}[t]{0.32\textwidth}
        \centering
        \includegraphics[width = \linewidth]{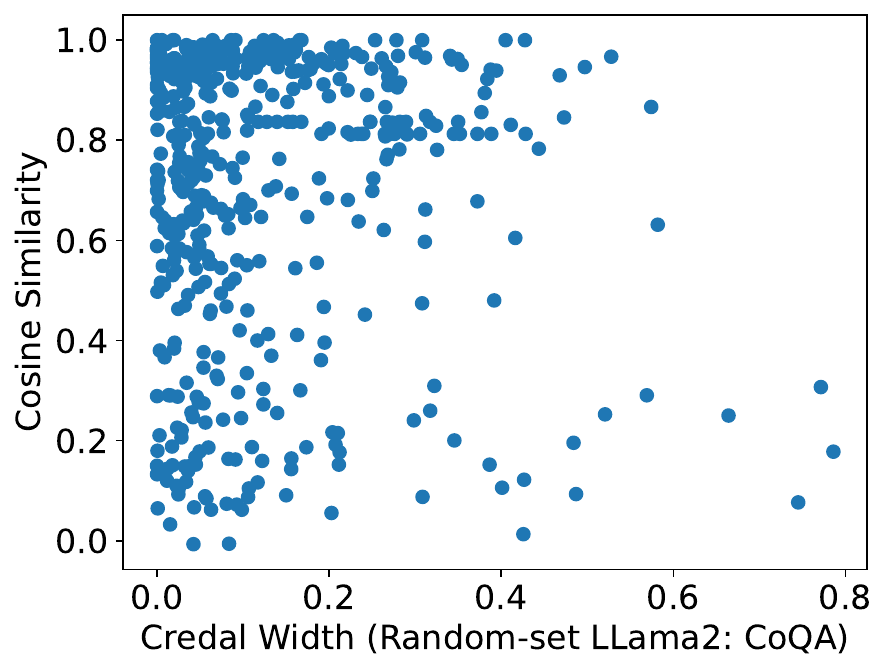}
        \caption{}
    \label{fig:Credal_width_CosSim_RS_LLama2_CoQA}
    \end{subfigure}%
    \\
    \begin{subfigure}[t]{0.32\textwidth}
        \centering
        \includegraphics[width = \linewidth]{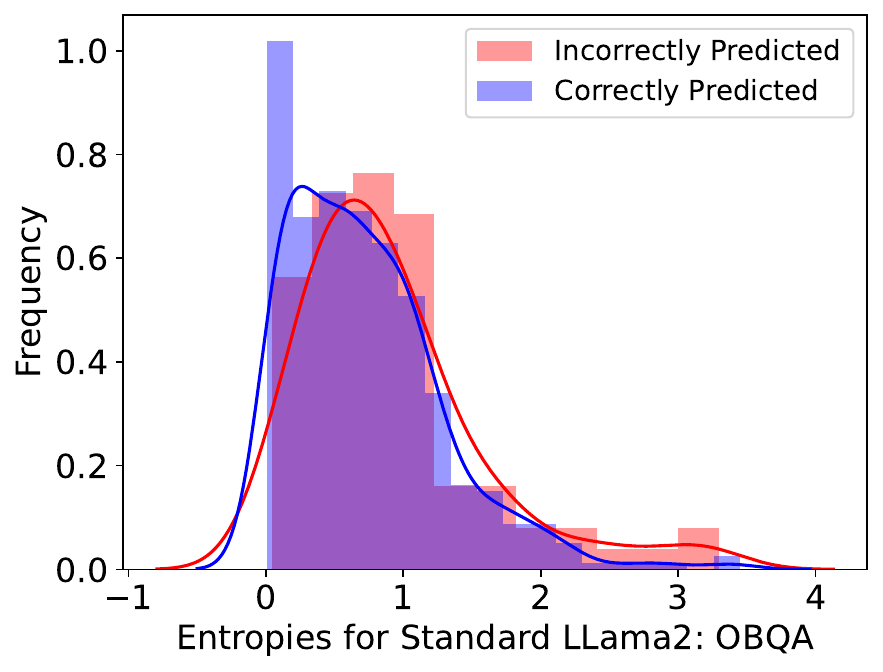}
        \caption{}
        \label{fig:Entropy_Standard_LLama2_OBQA}
    \end{subfigure}%
        \begin{subfigure}[t]{0.32\textwidth}
        \centering
        \includegraphics[width = \linewidth]{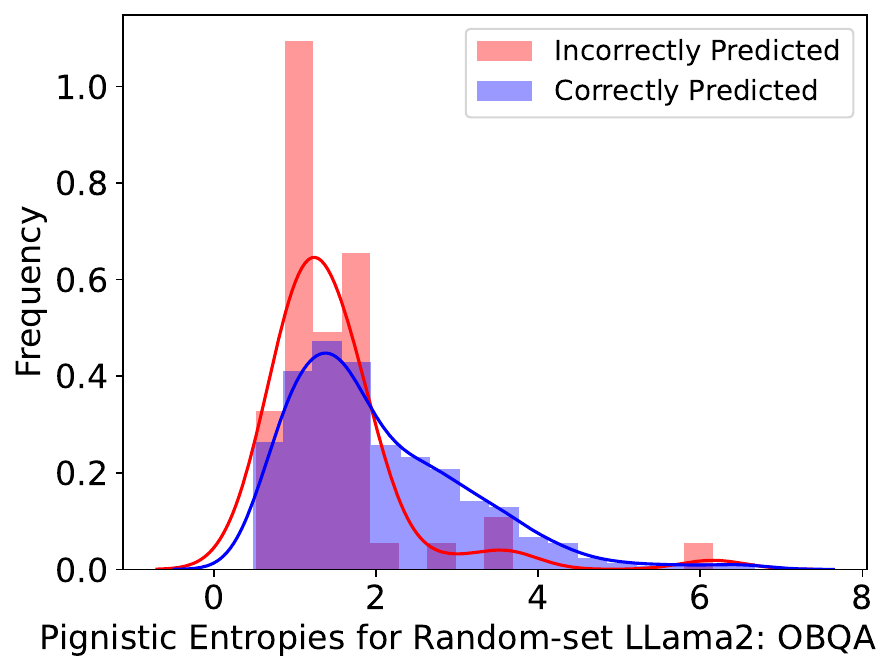}
        \caption{}
        \label{fig:Pig_Entropy_RS_LLama2_OBQA}
    \end{subfigure}%
    \begin{subfigure}[t]{0.32\textwidth}
        \centering
        % \vspace{1pt}
        \includegraphics[width = \linewidth]{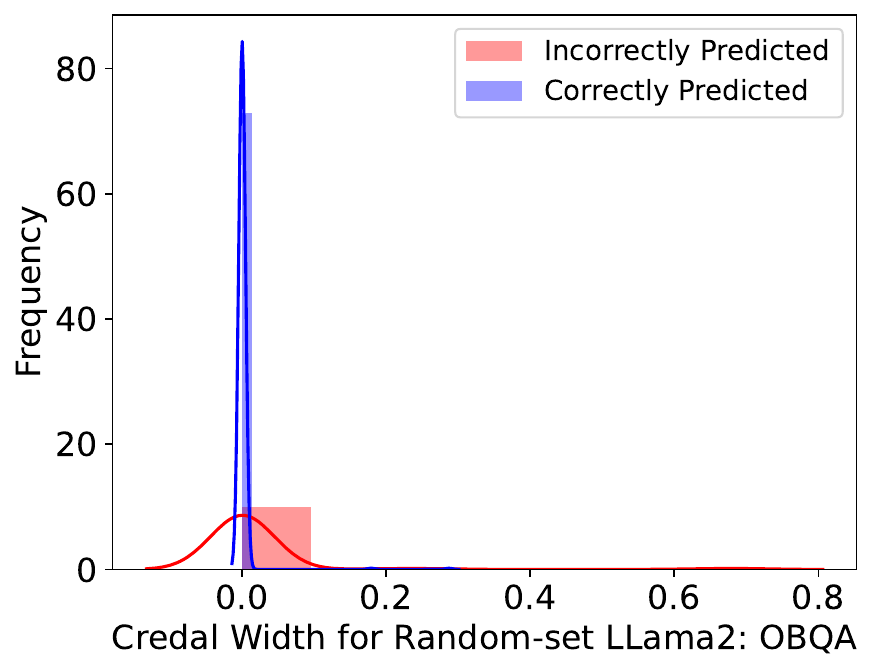}
        \caption{}
        \label{fig:Credal_width_RS_LLama2_OBQA}
    \end{subfigure}%
    \caption{Behavior of uncertainty measures of Llama2 and RS-Llama2 with respect to the correctness and closeness to the groundtruth on CoQA and OBQA datasets.}
    \label{fig:plots}
\end{figure*}

\begin{table}[h!]
    \caption{Performance of Standard LLMs and RS-LLMs on CoQA and OBQA datasets. For CoQA, cosine similarity is reported while accuracy is reported for OBQA.}
    \centering
    \begin{tabular}{|c|c|c|c|}
        \hline
         & Metric & CoQA & OBQA \\ 
        \hline
        \hline
        \multirow{2}{*}{Llama2} & \emph{Accuracy} & - & 83.20 \\ 
        \cline{3-4} 
        & \emph{Cosine Similarity} & 0.69 & - \\ 
        \cline{3-4} 
        % & \emph{Entropy} & 7.43 $\pm$ 0.93 & D2 \\ 
        % \cline{2-4} 
        % & B2 & C2 & D2\\ 
        \hline
        \multirow{2}{*}{RS-Llama2} & \emph{Accuracy} & - & \textbf{89.60}\\ 
        \cline{3-4} 
        & \emph{Cosine Similarity} & \textbf{0.71} & - \\ 
        % \cline{3-4} 
        % & \emph{Pignistic Entropy} & C2 & D2 \\ 
        % \cline{3-4} 
        % & \emph{Credal Width} & C2 & D2 \\  
        \hline
        \hline
        \multirow{2}{*}{Mistral} & \emph{Accuracy} & - & 91.60 \\ 
        \cline{3-4} 
        & \emph{Cosine Similarity} & 0.67 & - \\ 
        \cline{3-4} 
        % & \emph{Entropy} & 7.43 $\pm$ 0.93 & D2 \\ 
        % \cline{2-4} 
        % & B2 & C2 & D2\\ 
        \hline
        \multirow{2}{*}{RS-Mistral} & \emph{Accuracy} & - & \textbf{93.00}\\ 
        \cline{3-4} 
        & \emph{Cosine Similarity} & \textbf{0.72} & - \\ 
        % \cline{3-4} 
        % & \emph{Pignistic Entropy} & C2 & D2 \\ 
        % \cline{3-4} 
        % & \emph{Credal Width} & C2 & D2 \\  
        \hline
        \hline
        \multirow{2}{*}{Phi2} & \emph{Accuracy} & - & 87.60 \\ 
        \cline{3-4} 
        & \emph{Cosine Similarity} & 0.72 & - \\ 
        \cline{3-4} 
        % & \emph{Entropy} & 7.43 $\pm$ 0.93 & D2 \\ 
        % \cline{2-4} 
        % & B2 & C2 & D2\\ 
        \hline
        \multirow{2}{*}{RS-Phi2} & \emph{Accuracy} & - & \textbf{91.80}\\ 
        \cline{3-4} 
        & \emph{Cosine Similarity} & \textbf{0.73} & - \\ 
        % \cline{3-4} 
        % & \emph{Pignistic Entropy} & C2 & D2 \\ 
        % \cline{3-4} 
        % & \emph{Credal Width} & C2 & D2 \\  
        \hline
    \end{tabular}
    
    \label{tab:correctness_performance}
\end{table}

\begin{table*}[h]
    \caption{Uncertainty evaluation of Standard Llama2 and RS-Llama2 on correct and incorrect context.}
    \centering
    \begin{tabular}{|c|c|c|c|c|c|}
        \hline
         &  \multirow{2}{*}{Metric} & \multicolumn{2}{|c|}{Correct Context ($\downarrow$)} & \multicolumn{2}{|c|}{Inorrect Context($\uparrow$)} \\ 
        \cline{3-6} 
         & & CoQA & OBQA & CoQA & OBQA \\ 
        \hline
        \multirow{1}{*}{Llama2} & \emph{Entropy} & $0.39 \pm 0.45$ &  $0.75 \pm 0.59$ & $0.90 \pm 0.80$ &$1.10 \pm 0.93$ \\ 
        \cline{3-4} 
        % & \emph{Cosine Similarity} & 68.77 & - \\ 
        % \cline{3-4} 
        % & \emph{Entropy} & 7.43 $\pm$ 0.93 & D2 \\ 
        % \cline{2-4} 
        % & B2 & C2 & D2\\ 
        \hline
        % \multirow{2}{*}{RS-Llama2} & \emph{Pignistic Entropy} & $3.52 \pm 1.96$ & $1.97 \pm 1.09$  & $5.35 \pm 2.69$ & $1.61 \pm 0.94$\\ 
        % \cline{3-4} 
        % & \emph{Cosine Similarity} & 0.72 & - \\ 
        % \cline{3-6} 
        RS-Llama2 & \emph{Credal Width} & $0.13 \pm 0.13$ & $0.00 \pm 0.04$ & $0.28 \pm 0.21$ & $0.02 \pm 0.08$ \\  
        \hline
    \end{tabular}
    
    \label{tab:hallucination}
\end{table*}

\subsection{Results and Analysis}

\textbf{Performance comparison}.
As anticipated, we evaluate both models on CoQA and OBQA datasets. We employ greedy generation method for both models and datasets. For CoQA, we produce free text. To evaluate the closeness of generated and ground truth text, we employ the cosine similarity metric between the two \cite{banerjee2023benchmarking}. 
% {\color{red}Any reference?} 
Whereas, for OBQA, the option label is expected and generated which allows us to conveniently measure the accuracy w.r.t. the ground truth. Tab. \ref{tab:correctness_performance} reports the cosine similarity and accuracy on the CoQA and OBQA datasets, respectively, for both models. RS-LLMs clearly outperform the standard LLMs model on both datasets across all datasets, even though all models are trained using the exact same regimen. This clearly shows the representative prowess of random sets in the output space. Sample generations are shown in Appendix \ref{app:generated_results}.

\textbf{Uncertainty quantification}.
In Fig. \ref{fig:plots}, instead, we show how the uncertainty measures produced by the two models behave with respect to the correctness or closeness to the true answer. 
Figs. \ref{fig:Pig_Entropy_CosSim_RS_LLama2_CoQA} and \ref{fig:Entropy_CosSim_Standard_LLama2_CoQA} show the correlation between entropy and cosine similarities of RS-Llama2 and Standard Llama2, respectively, on the CoQA dataset. 
%{\color{cyan}
Pignistic entropy and cosine similarity for RS-Llama show some negative correlation 
whereas there is even lower correlation between entropy and cosine similarity for Llama2. 
% {\color{red}TBF, the negative trend is not so pronounced}, 

Figs. \ref{fig:Pig_Entropy_RS_LLama2_OBQA} and \ref{fig:Entropy_Standard_LLama2_OBQA} illustrate the correct vs. incorrect entropy distributions for RS-Llama2 and Llama2, respectively, on the OBQA dataset. 
Both models show no real trend here. 
% {\color{red}Results don't seem to favour us, do they?}
Ideally, the entropy distribution for correct predictions should be on the left of (away from) the entropy distribution of incorrect predictions. 
%{\color{cyan}
Lastly, the trend in Fig. \ref{fig:Credal_width_CosSim_RS_LLama2_CoQA} exhibit some negative correlation again between credal set width and cosine entropy. Fig. \ref{fig:Credal_width_RS_LLama2_OBQA} demonstrates the separability of correct and incorrect predictions using credal set width.

% {\color{red}In fact, there seems to be a bigger spread for incorrect predictions, right?}
We deliberate that this effect can be further enhanced by using a bigger budget as the chosen budget ($8,000$) is very small as compared to the total number of subsets possible ($2^{32000}$) so the distributions of masses is limited.
% So, it doesn't allow the masses to distribute well.
% We deliberate that this is because of the limited representation of focal sets in the budget. The chosen budget ($8,000$) is very small as compared to the total number of subsets possible ($2^{32000}$). 
% So, it doesn't allow the masses to distribute well. Therefore, using a bigger budget can solve this problem. 
% {\color{red}Sound weak, and imprecise. We should make a more rigorous argument I think, and possibly ablate the number of focal sets or their size}.

% {\color{red}Most of this argument seems to counter our claim of better uncertainty quantification. Consider what to keep. Also, these figures currently occupy an awful lot of space.}

\subsection{Hallucination}

Lastly, we experiment to evaluate the models on their hallucination. For this, we deliberately provide the LLM with incorrect context and then measure how confidently the model answers the given question. More specifically, for CoQA we randomly replace the question of given test sample with another question from the test sample and for OBQA we replace the answer choices with some other random choices. Tab. \ref{tab:hallucination} presents the uncertainty evaluation of Llama2 and RS-Llama2 under correct and incorrect context. %As evident, 

% {\color{cyan}
% RS-Llama2 exhibits good separability based on credal width for both datasets
% }
% {\color{red}In fact, Llama2 has lower entropy for incorrect answer on OBQA, so it arguably fails the hallucination test just like RS-Llama. Also, what counts is the relative separation, isn't it, not the absolute value of the entropy, correct? In this sense, the results are quite comparable.}
% {\color{cyan}sorry fabio, I accidentally flipped thenumbers for llama2. so it does spearate out better}
% {\color{cyan}
% whereas Llama2 too separates correct and incorrect contexts in terms of entropy. This shows that model is able to reason about second-level uncertainty using the credal width.
% }

Both models exhibit good separability on correct and incorrect contexts for both datasets. However, RS-Llama2 does so using credal width which better estimates the second-level uncertainty as detailed in Sec. \ref{sec:background}.

% {\color{red}Based on these results, should be talk about pignistic entropy at all? Also, there is no trace of the discussion we had about this high entropy being due to the spreading of masses among large focal sets of tokens.}

\section{Conclusions}
\label{sec:conclusions}

This paper proposes a novel training and fine-tuning regimen for Large Language Models (LLMs) that not only provides a higher-quality representation, leading to significantly improved performance, but also allows the model 
%the capability to live in a higher space than the usual probability space 
to encode second-order uncertainty on its own predictions, in the random set formalism.
%allowing the model to opportunity to reason about 2nd level uncertainty in its prediction. 
Through our experiments, we conclusively demonstrate the efficacy of our proposed approach. The results demonstrate the potential of our approach 
%presents an alternate 
to change the fundamental nature of LLMs, and move towards language models 
%which are better and more expressive about their uncertainty.
more aware of their own uncertainty and thus, potentially, less prone to hallucination.

A current limitation of our work is having to manually set the number of focal elements $K$; a dynamic strategy adjusting $K$ based on overlap would enhance flexibility and effectiveness and will be subject to future work.

Future work includes the exploration of other clustering algorithms (e.g., fuzzy clustering) for more effective budgeting, and the design of a large-scale mathematical framework for 
%to take it to much larger scale by 
training an LLM from scratch in the Random-Set approach, as our next objective. Another interesting line of investigation could be exploring the semantic relation of tokens in a focal set, to potentially propose random-sets as alternative to individual tokens altogether.

\bibliography{bibliography}
\bibliographystyle{icml2025}

%%%%%%%%%%%%%%%%%%%%%%%%%%%%%%%%%%%%%%%%%%%%%%%%%%%%%%%%%%%%%%%%%%%%%%%%%%%%%%%
%%%%%%%%%%%%%%%%%%%%%%%%%%%%%%%%%%%%%%%%%%%%%%%%%%%%%%%%%%%%%%%%%%%%%%%%%%%%%%%
% APPENDIX
%%%%%%%%%%%%%%%%%%%%%%%%%%%%%%%%%%%%%%%%%%%%%%%%%%%%%%%%%%%%%%%%%%%%%%%%%%%%%%%
%%%%%%%%%%%%%%%%%%%%%%%%%%%%%%%%%%%%%%%%%%%%%%%%%%%%%%%%%%%%%%%%%%%%%%%%%%%%%%%
\newpage
\appendix
\onecolumn
% \section{You \emph{can} have an appendix here.}

% You can have as much text here as you want. The main body must be at most $8$ pages long.
% For the final version, one more page can be added.
% If you want, you can use an appendix like this one.  

% The $\mathtt{\backslash onecolumn}$ command above can be kept in place if you prefer a one-column appendix, or can be removed if you prefer a two-column appendix.  Apart from this possible change, the style (font size, spacing, margins, page numbering, etc.) should be kept the same as the main body.

\section{Budgeting analysis}\label{sec:budgeting-analysis}
In this section, we perform an analysis of the budgeted focal sets obtained using the proposed budgeting technique. For this analysis, we utilize Llama2-7b-hf having a token size of 32000 and set $K-8000$ (i.e. we budget 8000 focal sets from $2^{32000}$).

\begin{figure*}[h]
    \centering
        \begin{subfigure}[t]{0.5\textwidth}
        \centering
        \includegraphics[width = \linewidth]{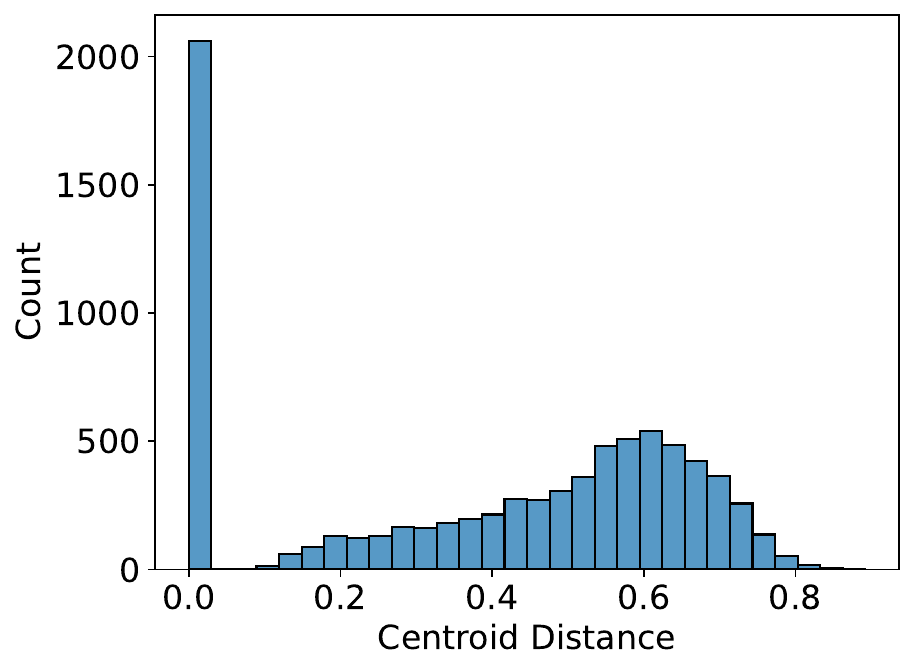}
        \caption{}
        \label{fig:centroid_distances}
    \end{subfigure}%
    \begin{subfigure}[t]{0.5\textwidth}
        \centering
        \includegraphics[width = \linewidth]{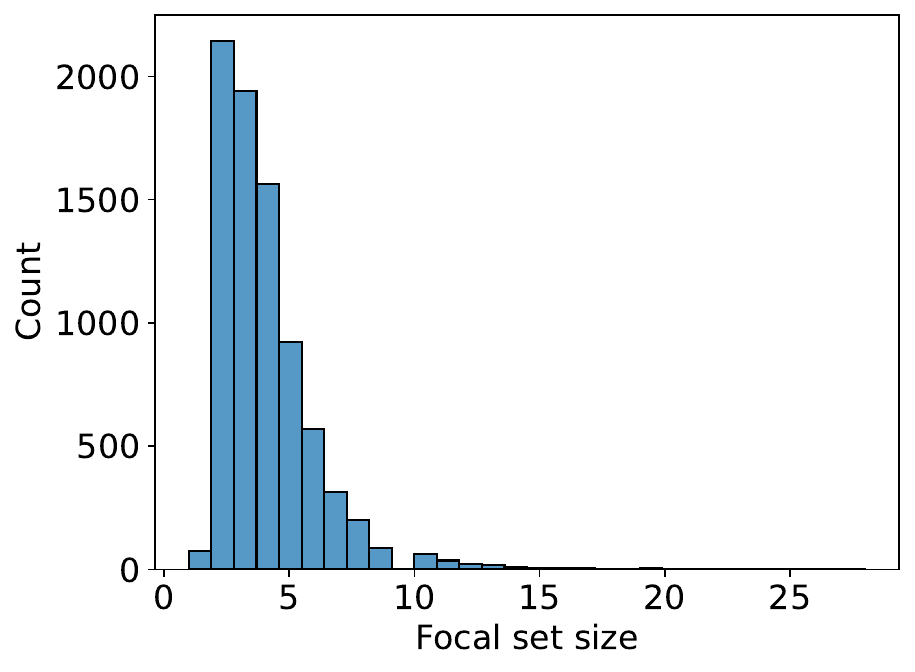}
        \caption{}
        \label{fig:focal_set_size}
    \end{subfigure}%       
    \caption{(a) Frequency distribution of centroid distances of the obtained budgeted focal sets. (b) Frequency distribution of sizes of obtained budgeted focal sets. Note: There are 8 focal sets with sizes $>30$. They are excluded here for better visualisation.}
    \label{fig:budgeting_analysis}
\end{figure*}

The obtained focal sets, in general, show appropriate closeness. Some examples include '[\verb+cattle+, \verb+ sheep+]', '[\verb+shame+, \verb+ pity+]', '[\verb+quiet+, \verb+ calm+, \verb+ quietly+]', '[\verb+maintain+, \verb+ retain+, \verb+ maintained+, \verb+ retained+]' and '[\verb+delight+,  \verb+ pleasure+, \verb+ pleased+, \verb+ proud+, \verb+ pride+]'. Furthermore, we perform semantic analysis of sets by computing centroid distance (mean euclidean distance of the elements of a cluster from its center) of each the sets. This distance is computer over the encoding obtained using sentence-transformers \cite{sentencetransformers} of each of the token. Fig. \ref{fig:centroid_distances} shows the distribution of centroid distances. Majority of them lie very close to zero with only away from it. This shows that the budgeting method does generate semantically meaningful sets.

Fig. \ref{fig:focal_set_size} shows the frequency count of the focal set sizes thus obtained. Most of the focal sets have a small size with 2,3 and 4 being the top most common sizes with counts of 2142, 1940 and 1562 respectively. The largest set size is 162 and it contains tokens like punctuations and symbols (., !, \#, \$, \%, etc).

\section{Ablation Studies}

\subsection{Hyperparameters $\alpha \: \& \: \beta$}
$M_s$ and $M_r$, defined in Eqs. \ref{eq:mass-reg} and \ref{eq:mass-sum} respectively, in the loss function (Eq. \ref{final_loss}) encourage that the model generates valid belief functions by constraining the sum of Masses to 1 and ensuring non-negativity of masses. However, there must be a balance on the amount of weight these terms should carry, as excessive penalization of deviations may hinder the model’s
ability to predict accurately. For example, in a Variational Auto Encoder (VAE), if we
assign too much weight to the KL divergence term in the loss, the model may prioritize fitting the
latent distribution at the expense of reconstructing the input data accurately. This imbalance can
lead to poor reconstruction quality and suboptimal performance on downstream tasks.

Tab. \ref{tab:ablation_hyperparameters} shows the cosine similarities using different values of $\alpha$ and $\beta$ for RS-Llama2. We observe that $\alpha=\beta=0.01$ and $\alpha=\beta=0.0001$ produce similar results. However, to encourge valid belief functions more, we use $\alpha=\beta=0.01$ in our experiments.

\begin{table}[h]
\centering
\begin{tabular}{|c|c|c|c|c|}
\hline
$\alpha = \beta$ & 0.1 & 0.01 & 0.001 & 0.0001 \\
\hline
Cosine Similarity & 0.66 & \textbf{0.71} & 0.69 & \textbf{0.71} \\
\hline
\end{tabular}
\caption{Cosine Similarity values for different $\alpha = \beta$ settings}
\label{tab:ablation_hyperparameters}
\end{table}

% $M_s$ and $M_r$, defined in Eqs. \ref{eq:mass-reg, eq:mass-sum}, in the loss function (Eq. \ref{final_loss}) encourage that the model generates valid belief functions by constraining the sum of Masses to 1 and ensuring non-negativity of masses. However, there must be a balance on the amount of weight these terms should carry, as excessive penalization of deviations may hinder the model’s
% ability to predict accurately.

\subsection{Focal Sets}
The number of non-singleton focal sets K to be budgeted is a hyperparameter and needs to be
studied. A lower value of K can lead to results more similar to those of classical LLMs, while a higher value of K can increase the complexity. Therefore, we conducted an ablation study on K on the CoQA dataset (Tab. \ref{tab:budget_cosine_similarity}). "Combined" represents the union of 2000, 4000, 8000, and 16000 budgets. We find that using a medium value for K (neither too small nor too big) leads the the most optimal results. These experiments are performed on Llama2 with $\alpha=\beta=0.01$.

\begin{table}[h]
\centering
\begin{tabular}{|c|c|c|c|c|c|}
\hline
Budget Size & 2000 & 4000 & 8000 & 16000 & Combined (56538) \\
\hline
Cosine Similarity & 0.67 & 0.68 & \textbf{0.71} & 0.66 & 0.69 \\
\hline
\end{tabular}
\caption{Cosine Similarity values across different budget sizes}
\label{tab:budget_cosine_similarity}
\end{table}

\section{Additional Experiments}
To further test our proposed approach, we evaluate our approach on a summarization task. We use the XSum \citep{narayan2018don} dataset with the task of summarizing a document in one line. Tab. \ref{tab:xsum_cosine_similarity} shows the performance of Mistral and RS-Mistral on the XSum dataset. RS-Mistral outperforms Standard Mistral and obtains better cosine similarity between the generated and ground truth summaries.

\begin{table}[h!]
\centering
\begin{tabular}{|c|c|c|}
\hline
 & Mistral & RS-Mistral \\
\hline
Cosine Similarity & 0.56 & \textbf{0.60} \\
\hline
\end{tabular}
\caption{Performance of Standard Mistral and RS-Mistral on XSum dataset.}
\label{tab:xsum_cosine_similarity}
\end{table}

\section{Generated Results} \label{app:generated_results}
In this section, we present the generations by RS-Llama2 and Llama2. Tab. \ref{tab:coqa_results} displays the their generations on the CoQA test set while Tab. \ref{tab:obqa_results} displays the generated results on the OBQA test set. For better readability, the 'story' in Tab. \ref{tab:coqa_results} has been shortened.

\begin{table}[t]
    \caption{Generated results by RS-Llama2 and Llama2 on CoQA test set.}
    \centering
    \begin{tabular}{|p{3cm}|p{3cm}|p{3cm}|p{3cm}|p{3cm}|}
\hline
Story & Question & RS-Llama2 & Llama2 & Groundtruth \\
\hline
The Six-Day War (Hebrew: , "Milhemet Sheshet Ha Yamim"; Arabic: , "an-Naksah", " $\dots$ ing to the conflict. & When was the Six-Day War fought? & 1967. & 1967. & June 5 and 10, 1967 \\ 
\hline
(CNN) -- President Barack Obama has nominated his top counterterrorism adviser,  $\dots$ the past year or so. & Who did Obama nominate? &  John Brennan, to be the next director of the CIA. &  John Brennan. & John Brennan. \\ 
\hline
(CNN) -- There are plenty of country songs about how to be a man. One way is to  $\dots$ oses at him over it. & Who came out? & 2 country crooners. &  Ty Herndon and Billy Gilman came out as being gay. & Ty Herndon and Billy Gilman \\ 
\hline
There once was a little old lady who had no friends. She rode her bike to and fr $\dots$ nd her first friend. & Who was antosocial at first? &  the little old lady. &  the little old lady. & the old lady \\ 
\hline
Weimar Republic was an unofficial, historical designation for the German state b $\dots$  the Eastern border. & What was an unofficial designation & 1919 and 1933. &  Weimar Republic  It was first used in a book by Wolfgang Mommsen in 1983. & the Weimar Republic \\ 
\hline
CHAPTER XX A BLIZZARD A bitter wind searched the poplar bluff where George and h $\dots$  the withered brush. & Who were they waiting for? &  Edgar, who should have joined them with the team and sledge. &  West and the mail-carrier. & Edgar \\ 
\hline
There was a princess named Janet who was very pretty. She wanted to marry a prin $\dots$ after in the castle. & Who was the princess? &  Janet. &  Janet. & Janet \\ 
\hline
Jenny's family lived in a small apartment in Seattle. One day Jenny came home fr $\dots$  going to like Utah! & Did Jenny help her mom pack? &  yes. &  Yes. & yes \\ 
\hline
\end{tabular}
    
    \label{tab:coqa_results}
\end{table}

\begin{table}[t]
    \caption{Generated results by RS-Llama2 and Llama2 on OBQA test set.}
    \centering
    \begin{tabular}{|p{3cm}|p{5cm}|p{2cm}|p{2cm}|p{2cm}|}
\hline
Fact & Question & RS-Llama2 & Llama2 & Groundtruth \\
\hline
ugars are transported from the leaves to the roots of a plan & Glucose travels
A) from roots to leaves of a daffodil
B) from a rose's leaves to the atmosphere
C) from a daisy's leaves into it's underground support system
D) from the sun to a sunflower's buds
 &  C &  C & C \\ 
\hline
he use of alternative fuels increases , the use of gasoline will decrease & As gasoline costs rise, alternative fuels are being used, which means that
A) wind power will be expensive
B) gas costs will rise
C) oil costs will be maintained
D) gasoline will be needed less
 &  D &  D & D \\ 
\hline
he Grand Canyon was formed by the Colorado River flowing over long periods of time & The Grand Canyon was formed by
A) a volcano erupting in 1782
B) a river named after the 20th state to join the union flowing over time
C) a river named after the 38th state to join the union flowing over time
D) the Great Lakes drying up
 &  B &  B & C \\ 
\hline
orms cause bodies of water to increase amount of water they contain & After a torrential downpour over a week, a man notices that the pond in his backyard is
A) melted
B) dehydrated
C) bloated
D) salted
 &  C & C 
 & C \\ 
\hline
hick feathers can be used for keeping warm & What can feathers on Spheniscidae be used for?
A) keeping warm
B) flying
C) sleeping
D) eating
 &  A &  A & A \\ 
\hline
decomposition increases the amount of nutrients in the soil & They looked where the log decayed to garden as it would leave the earth
A) richer
B) dryer
C) sandy
D) harder
 &  A &  A & A \\ 
\hline
plant requires soil for to grow & A plant will grow strong if it has
A) love
B) heat
C) earth
D) sand
 &  C & C
 & C \\ 
\hline
ometimes weathering smooths rock & Over a period of time the weather can change
A) The color of my hair
B) The way I walk
C) The size of a statue
D) The sound a computer makes
 &  C &  C & C \\ 
\hline
wind and rain cause erosion & Erosion is caused by different kinds of
A) soil
B) fish
C) rocks
D) weather
 &  D &  D & D \\ 
\hline
photosynthesis makes energy for the plant by converting carbon dioxide, water, and sunlight into carbohydrate & Which of the following is not an input in photosynthesis?
A) sunlight
B) oxygen
C) water
D) carbon dioxide
 &  B &  B & B \\ 
\hline
\end{tabular}
    
    \label{tab:obqa_results}
\end{table}

%%%%%%%%%%%%%%%%%%%%%%%%%%%%%%%%%%%%%%%%%%%%%%%%%%%%%%%%%%%%%%%%%%%%%%%%%%%%%%%
%%%%%%%%%%%%%%%%%%%%%%%%%%%%%%%%%%%%%%%%%%%%%%%%%%%%%%%%%%%%%%%%%%%%%%%%%%%%%%%

\end{document}